\documentclass{article}

\usepackage{microtype}
\usepackage{graphicx}
\usepackage{subcaption}
\usepackage{multirow}
\usepackage{booktabs} 

\usepackage{hyperref}

\usepackage[preprint]{icml2026}

\usepackage{amsmath}
\usepackage{amssymb}
\usepackage{mathtools}
\usepackage{amsthm}

\usepackage[capitalize,noabbrev]{cleveref}

\theoremstyle{plain}

\theoremstyle{definition}

\theoremstyle{remark}

\usepackage[textsize=tiny]{todonotes}

\icmltitlerunning{TIDE: Text-Informed Dynamic Extrapolation with Step-Aware Temperature Control for Diffusion Transformers}

\begin{document}

\twocolumn[
  \icmltitle{TIDE: Text-Informed Dynamic Extrapolation with Step-Aware Temperature Control for Diffusion Transformers}



  \icmlsetsymbol{equal}{*}

  \begin{icmlauthorlist}
    \icmlauthor{Yihua Liu}{indep}
    \icmlauthor{Fanjiang Ye}{rice}
    \icmlauthor{Bowen Lin}{uh}
    \icmlauthor{Rongyu Fang}{indep}
    \icmlauthor{Chengming Zhang}{uh}

  \end{icmlauthorlist}

  \icmlaffiliation{indep}{Independent Researcher}
  \icmlaffiliation{uh}{University of Houston}
  \icmlaffiliation{rice}{Rice University}

  \icmlcorrespondingauthor{Yihua Liu}{fwcnigo@gmail.com}
  \icmlcorrespondingauthor{Chengming Zhang}{czhang48@uh.edu}

  \icmlkeywords{Machine Learning, Diffusion Transformer, Image Generation, Text-to-Image, Resoulution Extrapolation, ICML}

  \vskip 0.3in
]



\printAffiliationsAndNotice{}  

\begin{abstract}
 Diffusion Transformer (DiT) faces challenges when generating images with higher resolution compared at training resolution, causing especially structural degradation due to attention dilution.
Previous approaches attempt to mitigate this by sharpening attention distributions, but fail to preserve fine-grained semantic details and introduce obvious artifacts. 
In this work, we analyze the characteristics of DiTs and propose TIDE, a training-free text-to-image (T2I) extrapolation method that enables generation with arbitrary resolution and aspect ratio without additional sampling overhead.
We identify the core factor for prompt information loss, and introduce a text anchoring mechanism to correct the imbalance between text and image tokens. 
To further eliminate artifacts, we design a dynamic temperature control mechanism that leverages the pattern of spectral progression in the diffusion process. 
Extensive evaluations demonstrate that TIDE delivers high-quality resolution extrapolation capability and integrates seamlessly with existing state-of-the-art methods.
\end{abstract}
\begin{figure*}[t]
    \centering
    \includegraphics[width=\linewidth]{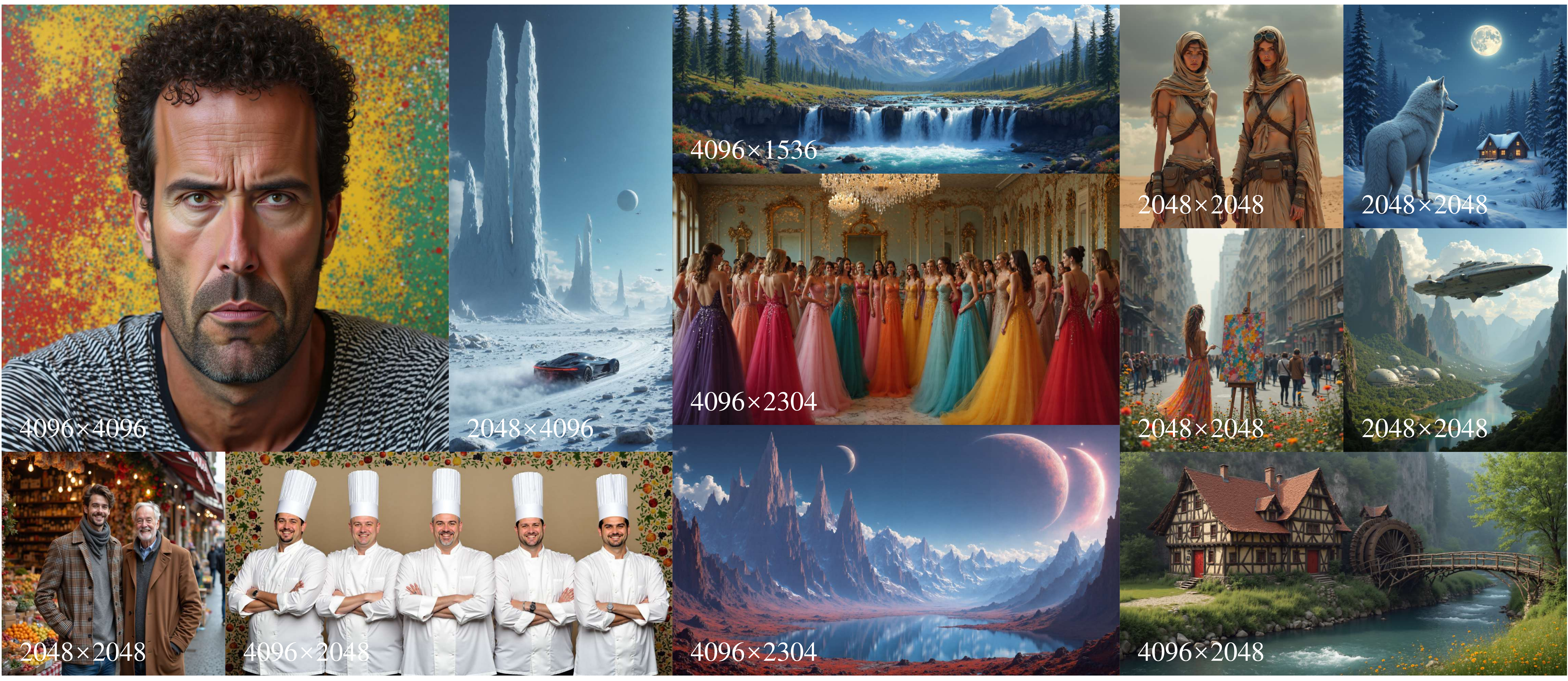} 
    
    \caption{Collage of multi-resolution results generated by TIDE. Prompts are from DrawBench and Aesthetic-4K. Zoom in for details.}
    \label{fig:collage}
\end{figure*}

\begin{figure*}[t] 
    \centering

    \includegraphics[width=0.9\textwidth]{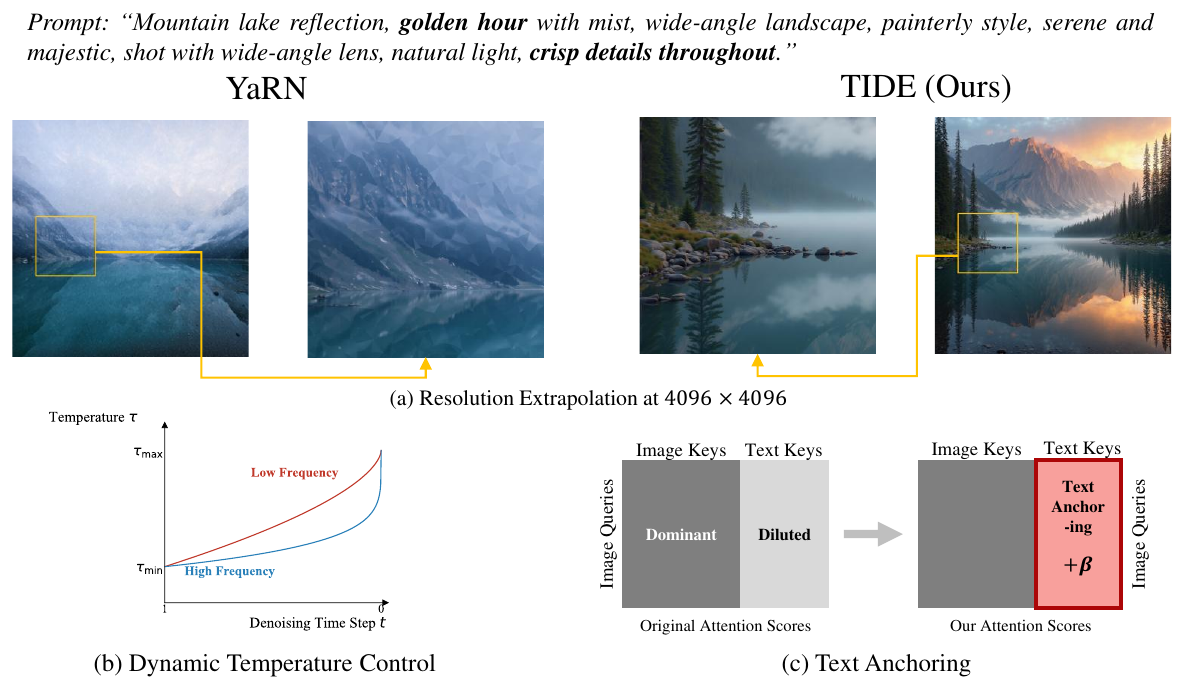}

    \caption{\textbf{Overview of our proposed TIDE framework.} (a) Visual comparison at $4096 \times 4096$ resolution. While existing methods (left) suffer from information loss and repetitive artifacts, our approach (right) preserves prompt fidelity and generates realistic details. (b) \textbf{Dynamic Temperature Control} method, which dynamically adjusts the attention temperature across time-steps to eliminate high-frequency artifacts. (c) \textbf{Text Anchoring} mechanism, which counteracts attention dilution by reinforcing the cross-attention scores between image queries and text keys, recovering the influence of text tokens.}
    
    \label{fig:overview} 
\end{figure*}
\section{Introduction}

Diffusion models \citep{sohl2015deep, ho2020denoising, ho2022classifier, ramesh2022hierarchical, rombach2022high} have shown impressive performance in text-to-image (T2I) generations. With its superior scalability, Diffusion Transformer (DiT) \citep{peebles2023scalable} has replaced UNet as the most popular architecture in recent state-of-the-art pre-trained diffusion models, such as Stable Diffusion 3 \citep{esser2024scaling} and FLUX \citep{flux2024}. However, generating images with higher resolution than pretrained resolution still leads to collapse in global structures and local details \citep{bu2025hiflow}. 
One possible solution is to train or fine-tune models directly at the higher resolution \citep{hoogeboom2023simple, xie2023difffit}. However, high-resolution image datasets are not as easily accessible as long text datasets. Moreover, DiT's sequence length has a quadratic relationship with the image token length, which increases demands on computational resources and memory. 
Another intuitive approach, which is more practical, is to apply super-resolution (SR) methods to low-resolution images \citep{wang2024exploiting, yang2024inf, duan2025dit4sr}. However, SR methods are fundamentally constrained by the low-resolution structural priors. They faithfully follow the low-resolution input and fail to generate novel semantic content or expand the field of view, which are necessary for creating a panoramic image with complex multi-object interactions. 
Therefore, a training-free extrapolation method is crucial for democratizing high-resolution synthesis.

Research on training-free resolution extrapolation for diffusion models can be broadly classified into two categories. The first category focuses on designing novel sampling strategies. Techniques proposed by \citet{bar2023multidiffusion, lin2024cutdiffusion} decompose the image into patches and maintain global coherence through region overlap or pixel relocation.  \citet{du2024demofusion, zhang2024frecas, haji2024elasticdiffusion, kim2025diffusehigh, bu2025hiflow, sanjyal2025rectifiedhr} utilized the cascaded sampling technique, where a high-resolution image is progressively generated starting from lower resolutions.
While these methods are effective and independent of specific model architectures, their complex sampling procedures introduce significant inference latency, preventing further system-level optimization.
The second category aims to modify the model architecture, enabling it to handle larger resolutions directly. Previous works for UNet \citep{he2023scalecrafter, lin2025accdiffusion} focused on expanding the receptive field with dilated convolutional layers. As DiT has replaced UNet as the primary backbone for diffusion models, research into DiT's extrapolation capabilities has currently become the most critical direction.

Transformer-based architectures inherently face two fundamental challenges in sequence extrapolation: out-of-distribution (OOD) positional embeddings and attention dilution. Extensive research has been devoted to mitigating these issues in the Large Language Model (LLM) domain \citep{su2024roformer, kaiokendev, chen2023extending, blocntkaware, peng2023yarn, ding2024longrope, nakanishi2025scalable}. However, the transition from 1D texts to 2D images introduces distinct challenges for DiTs. First, the standard 2D Rotary Positional Embedding (RoPE) \citep{su2024roformer, lu2024fit, zhuo2024lumina} implies that positional indices scale linearly with resolution, whereas the sequence length grows quadratically. This inconsistency makes attention dilution significantly more critical than positional OOD. Second, in T2I tasks with mainstream Multimodal Diffusion Transformers (MM-DiT) architecture \citep{esser2024scaling}, the fixed number of text tokens is severely overwhelmed by the quadratic expansion of image tokens, leading to diminished semantic guidance at high resolutions.

However, previous approaches for DiTs mainly focused on positional OOD \citep{lu2024fit, zhuo2024lumina, issachar2025dype}, leaving the attention dilution issue under-explored. \citet{jin2023training} provided a valuable theoretical analysis of this phenomenon in DiTs, yet their proposed approach is methodologically analogous to the generic attention sharpening used in YaRN \citep{peng2023yarn}, which has already been widely applied in LLMs. Even with recent refinements such as head-specific adjustments \citep{zhao2025ultraimage}, these approaches essentially rely on heuristic entropy reduction that fails to exploit the characteristics for DiTs and T2I tasks. Consequently, they suffer from two limitations: First, naively sharpening the distribution recovers only a few dominant tokens while suppressing other information. Second, employing a static sharpening strategy neglects the spectral pattern of the diffusion process \citep{rissanen2022generative, hoogeboom2023simple, issachar2025dype}, leading to severe high-frequency artifacts.

To bridge the gap between the complex attention distribution issue of DiTs and the simple sharpening technique, we propose TIDE, a training-free framework for high-resolution T2I generation. 
We first analyze the aggregate attention scores obtained by text tokens during the sampling process at different resolutions, confirming that the text influence significantly decays as resolution increases. 
Furthermore, we confirm that simply sharpening the attention distribution cannot completely resolve this issue and may instead introduce severe artifacts. 
Building on these insights, we provide two core innovations (illustrated in Figure~\ref{fig:overview}). \textbf{First}, we introduce Text Anchoring to restore the influence of text tokens during the sampling process and improve the quality of global structure. \textbf{Second}, we propose a Dynamic Temperature Control scheme based on the pattern of spectral progression in the diffusion process, which effectively eliminates the high-frequency local artifacts commonly caused by static sharpening strategies. Overall, our approach unlocks the full potential of DiTs' capabilities for generation at arbitrary resolutions and aspect ratios, ensuring that the model's architectural flexibility does not come at the cost of image quality degradation.

Our main contributions are summarized as follows:
\vspace{-0.5em}
\begin{itemize}
    \setlength{\itemsep}{0pt}
    \setlength{\parsep}{0pt}
    \setlength{\parskip}{0pt}

    \item We analyze the issue of attention distribution in high-resolution synthesis with DiTs, which has been overlooked compared with the OOD issue of positional embeddings. We point out that current attention sharpening techniques are insufficient for solving this issue.
    
    \item We propose TIDE, a training-free framework without additional sampling steps, providing a solid foundation for general system-level inference optimizations.

    \item We perform extensive experiments and ablation studies, demonstrating the effectiveness of TIDE.

    \item Our framework can be seamlessly integrated with existing positional embeddings interpolation methods and sampling-based strategies.
    
\end{itemize}
\vspace{-0.5em} 

\section{Background}

\subsection{Diffusion Transformer for Text-to-Image Synthesis} \label{sec:dit_for_t2i}

Diffusion Transformer (DiT) \citep{peebles2023scalable} has replaced the U-Net backbone with a transformer that operates on latent space~\citep{rombach2022high} . Due to its flexible and scalable architecture, DiT has become the foundation of many state-of-the-art image generation models, such as Stable Diffusion 3 \citep{esser2024scaling}, FLUX.1 \citep{flux2024} and Qwen-Image \citep{wu2025qwen}. 

Previous DiT-based text-to-image synthesis methods \citep{chen2023pixart} mainly utilize cross-attention mechanisms to incorporate textual guidance. Formally, let $\mathbf{X}_{\text{I}}\in\mathbb R^{L_{\text{I}} \times c}$  and $\mathbf{X}_{\text{T}}\in\mathbb R^{L_{\text{T}} \times c}$ denote the features of the image token sequence and text token sequence, where $L_{\text{I}}$ , $L_{\text{T}}$ and $c$ are the length of image token sequence, the length of text token sequence and the token dimension respectively. Then we can define the key, query and value matrices as follows:

\begin{equation} \small 
    \mathbf{Q}_{\text{I}} = \mathbf{X}_{\text{I}}\mathbf{W}_Q, \mathbf{K}_{\text{T}}=\mathbf{X}_{\text{T}}\mathbf{W}_K, \mathbf{V}_{\text{T}}=\mathbf{X}_{\text{T}}\mathbf{W}_V
    \label{eq:cross_qkv}
\end{equation}

where $ \mathbf{W}_{ \{Q,K,V\} } \in \mathbb R^{c \times d}$ are learnable projection matrices, and $d$ is the hidden dimension. Attention is computed as:

\begin{equation} \small 
     \mathbf{S}= \mathbf{Q}_{\text{I}} \mathbf{K}_{\text{T}}^\top,  \mathbf{P}=\text{softmax}(\frac{ \mathbf{S}}{\sqrt{d}}),  \mathbf{O}= \mathbf{P} \mathbf{V}_{\text{T}}
    \label{eq:cross_attention}
\end{equation}

However, most current state-of-the-art models use the Multimodal Diffusion Transformer (MM-DiT) architecture from Stable Diffusion 3, where image tokens and text tokens are concatenated in the sequence length dimension before being processed by transformer blocks. Formally, we can define the input of transformer blocks as follows:

\begin{alignat}{2} \small 
     \mathbf{Q} &=\text{Concat}( \mathbf{Q}_{\text{T}}, \mathbf{Q}_{\text{I}}) &&=\text{Concat}( \mathbf{X}_{\text{T}} \mathbf{W}_{Q}^{\text{T}},  \mathbf{X}_{\text{I}} \mathbf{W}_{Q}^{\text{I}})  
     \label{eq:joint_q}\\
     \mathbf{K} &=\text{Concat}( \mathbf{K}_{\text{T}}, \mathbf{K}_{\text{I}}) &&=\text{Concat}( \mathbf{X}_{\text{T}} \mathbf{W}_{K}^{\text{T}},  \mathbf{X}_{\text{I}} \mathbf{W}_{K}^{\text{I}}) 
     \label{eq:joint_k}\\
     \mathbf{V} &=\text{Concat}( \mathbf{V}_{\text{T}}, \mathbf{V}_{\text{I}}) &&=\text{Concat}( \mathbf{X}_{\text{T}} \mathbf{W}_{V}^{\text{T}},  \mathbf{X}_{\text{I}} \mathbf{W}_{V}^{\text{I}})
    \label{eq:joint_v}
\end{alignat}

where $\text{Concat}(\cdot)$ denotes the concatenation operation along the sequence length dimension, and $ \mathbf{W}_{  \{Q,K,V\} }^{\text{I}} \in \mathbb R^{c \times d}$ and $ \mathbf{W}_{  \{Q,K,V\} }^{\text{T}} \in \mathbb R^{c \times d}$ denote the projection matrices for image features and text features. Joint attentions in MM-DiT can be computed as follows:

\begin{equation} \small 
     \mathbf{S}= \mathbf{Q} \mathbf{K}^\top, 
      \mathbf{P}=\text{softmax}(\frac{ \mathbf{S}}{\sqrt{d}}),
      \mathbf{O}= \mathbf{P} \mathbf{V}
    \label{eq:joint_attention}
\end{equation}

\subsection{Attention Entropy}
Since each row of the attention score matrix $ \mathbf{P}=[p_{ij}] \in \mathbb R^{L \times L}$  in Eq.~\eqref{eq:joint_attention} can be treated as a probability mass function of a discrete random variable, \citet{jin2023training} defined the attention entropy with respect to the token at position $i$ as follows:

\begin{equation} \small 
    H_i = -\sum_{j=1}^N p_{ij} \log(p_{ij})
    \label{eq:entropy_def}
\end{equation}

The entropy of the $i$-th token is maximized when attention scores are uniformly distributed and minimized when attention concentrates on a single token. Consequently, attention entropy quantifies the uniformity of the attention distribution. As demonstrated by \citet{jin2023training}, this entropy scales with the sequence length $L$:

\begin{equation} \small 
    H_{i} = \log L - \frac{\sigma_{i}^2}{2}
    \label{eq:entropy_growth}
\end{equation}

where  the latter item $\frac{\sigma_{i}^2}{2}$ is not relevant with the sequence length $L$.

Sequence extrapolation in Transformer-based architectures is fundamentally hindered by two challenges: out-of-distribution (OOD) positional embeddings and attention dilution. To address these concurrently, YaRN \citep{peng2023yarn} serves as a representative baseline, integrating a hybrid strategy that combines positional interpolation with attention distribution control.

\subsection{YaRN and Distribution Sharpening}

Transformer-based models, including both LLMs and DiTs, inherently face two fundamental challenges in sequence extrapolation: out-of-distribution (OOD) positional embeddings and attention dilution. YaRN \citep{peng2023yarn} serves as a representative baseline, integrating a hybrid strategy that combines positional interpolation with attention distribution control. Specifically, YaRN employs an NTK-by-parts interpolation scheme to align positional indices within the training domain, while introducing a temperature factor $\tau$ to the softmax operation to counteract attention dilution:
\begin{equation} \small 
    \mathbf{P} = \text{softmax}\left( \frac{\mathbf{Q}\mathbf{K}^\top}{\tau \sqrt{ d }} \right)
    \label{eq:attention_temperature}
\end{equation}
with extrapolation scaling factor $s$, the recommended temperature $\tau$ by YaRN is:

\begin{equation} \small 
    \sqrt{ \frac{1}{\tau} } = 0.1 \ln(s) + 1
    \label{eq:yarn_temperature}
\end{equation}

The $\tau$ value determined in Eq.~\eqref{eq:yarn_temperature} is always less than $1.0$ and decreases as $s$ increases. Consequently, the distribution of attention scores is sharpened, thereby mitigating the issue of attention dilution.
\section{Method}

\subsection{Analysis of Extrapolation Failure}

\begin{figure}[ht]
    \centering
    \includegraphics[width=\linewidth]{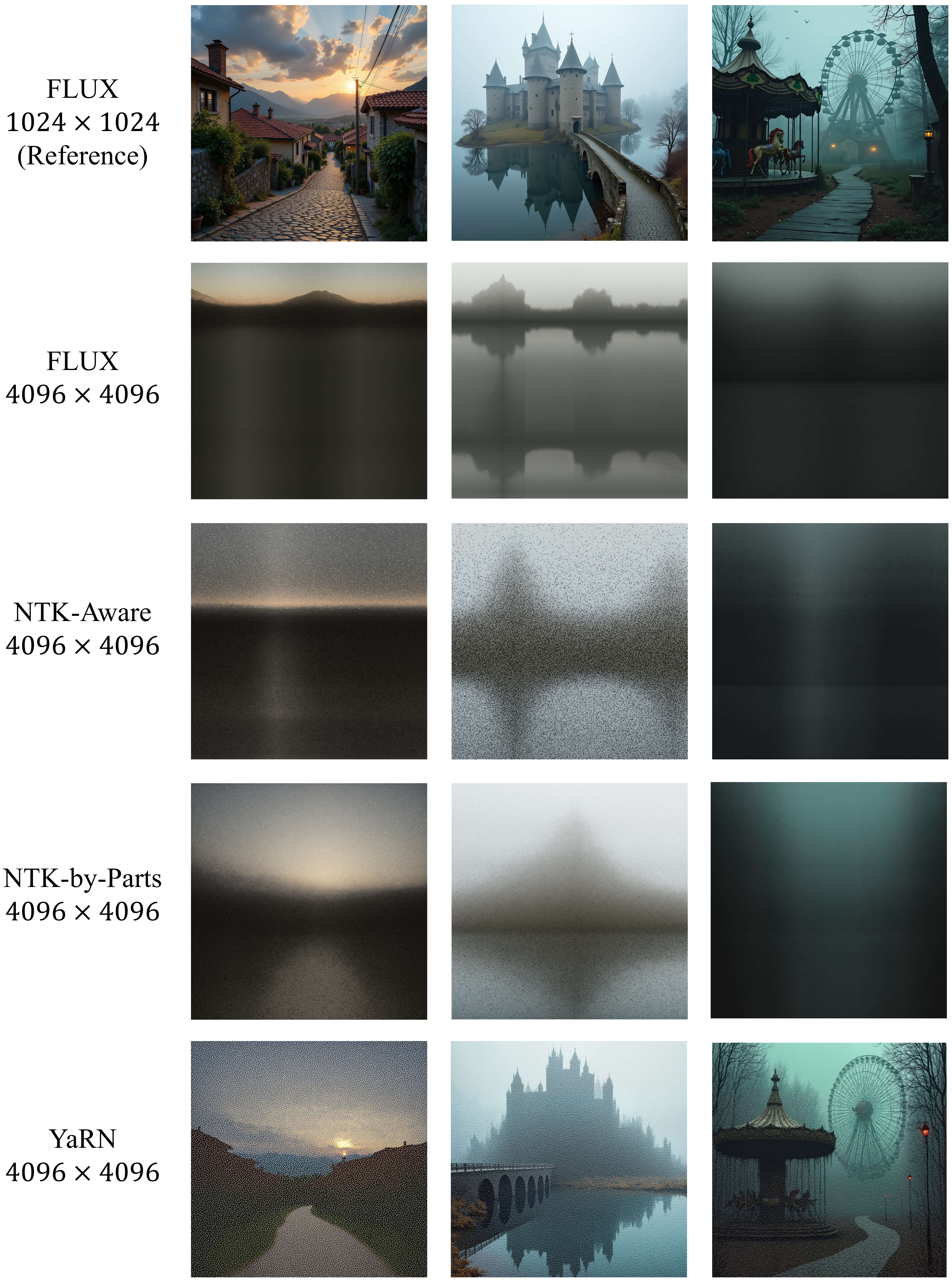}
    \caption{\textbf{Visual comparison of subject vanishing issues at 4K resolution.} We compare four extrapolation baselines at $4096 \times 4096$ with the native resolution $1024 \times 1024$. While the model at native resolution produces coherent subjects, baseline methods except YaRN suffer from severe subject vanishing issue, and YaRN's content richness has also noticeably declined.}
    \label{fig:subject_vanishing}
\end{figure}

\begin{figure*}[ht!]
    \centering
    \includegraphics[width=\textwidth]{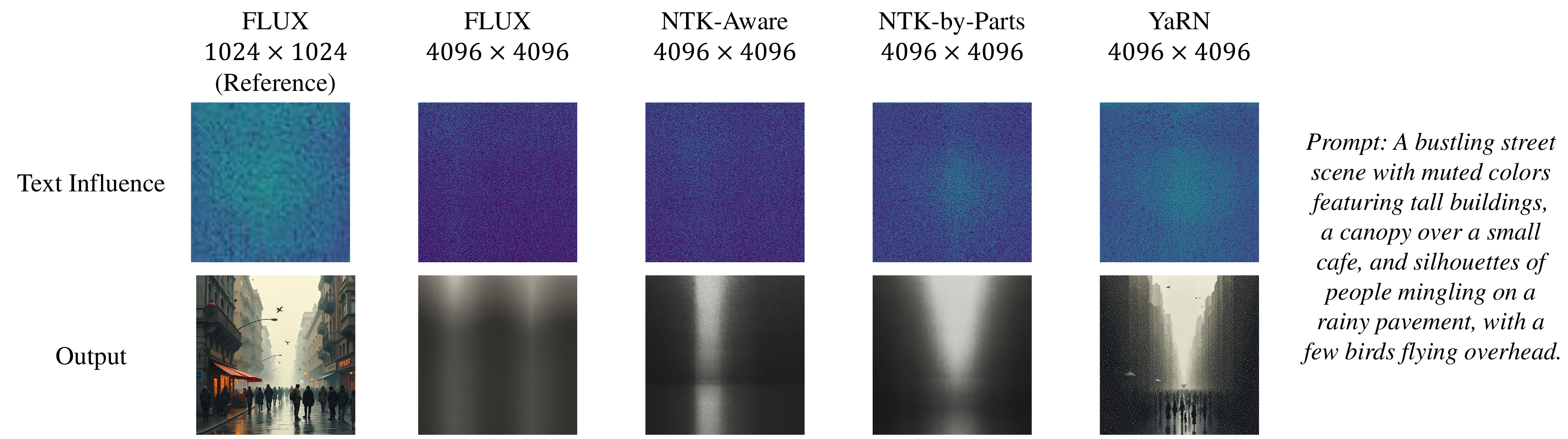}
    \caption{\textbf{Visualization of text token influence decay.} We visualize the spatial influence of text prompts at the early sampling stage, revealing that text guidance influence diminishes as the target resolution scales to $4096 \times 4096$. Specifically, Direct Extrapolation, NTK-Aware Interpolation and NTK-by-Parts Interpolation exhibit a near-total loss of spatial influence, providing a mechanistic explanation for the subject vanishing issue. YaRN only partially recovers token influence in the central region, resulting in suboptimal generation quality.}
    \label{fig:text_decay}
\end{figure*}
Equation~\eqref{eq:entropy_growth} suggests that as sequence length $N$ increases, the model suffers from attention dilution, where the distribution of attention score becomes overly smoothed. When generating high-resolution (e.g. $4096 \times 4096$) images with DiT, the distribution of attention scores can be overly uniform which implies that the output token is essentially the average of the value tokens. 

As illustrated in Figure~\ref{fig:subject_vanishing}, most extrapolation methods suffer from the issue of subject vanishing, because all tokens blend together after transformer blocks, ultimately resulting in a featureless image only having the average color. Only YaRN \citep{peng2023yarn} can mitigate this issue, yet it still cannot recover other information in the prompt as shown in reference images. Merely applying NTK-by-parts methods and removing the temperature factor in Eq.~\eqref{eq:attention_temperature} fails to solve this issue. This contrast confirms that it is the temperature $\tau$ defined in Eq.~\eqref{eq:yarn_temperature} that sharpens the distribution of attention score and reduces the attention entropy, therefore mitigates the issue of subject vanishing. Methods proposed by \citet{jin2023training, chiang2022overcoming, nakanishi2025scalable} essentially lower the temperature as well, which are not based on NTK-by-parts method, demonstrating that adjusting the attention distribution does not rely on specific positional embeddings interpolation methods.

Although lowering the temperature globally has noticeable results, this method originally designed for LLMs does not consider the characteristics of text-to-image generation with DiTs, which are text decay in high-resolution generation and the spectral progression of the diffusion process.

\noindent \textbf{Text Decay in High-Resolution Generation.}
As discussed in Section~\ref{sec:dit_for_t2i}, the image token sequence and text token sequence are usually concatenated to one sequence before computing attention. As image resolution grows, the length of image token sequence $L_{\text{I}}$ grows as well, while the length of text token sequence $L_{\text{T}}$ remains unchanged. This leads to the attention for the text token being severely diluted.

With $\mathbf{K}_{\text{T}} \in \mathbb R^{L_{\text{T}}\times d}$  and $\mathbf{K}_{\text{I}} \in \mathbb R^{L_{\text{I}}\times d}$ defined in Eq.~\eqref{eq:joint_k}, the attention score $\mathbf{P}$ in Eq.~\eqref{eq:joint_attention} can be rewritten as:

\begin{equation} \small
    \begin{split}
        \mathbf{P} &= \text{Concat}(\mathbf{P}_{\text{T}}, \mathbf{P}_\text{I}) \\
        &=\text{Softmax}\left(\frac{\text{Concat}(\mathbf{S}_{\text{T}}, \mathbf{S}_{\text{I}})}{\sqrt{d}}\right)
        \label{eq:concat_score}
    \end{split}
\end{equation}

where $\mathbf{S}_{\text{T}} = \mathbf{Q}\mathbf{K}_{\text{T}}^\top \in \mathbb R^{L \times L_{\text{T}}}$ and $\mathbf{S}_{\text{I}} = \mathbf{Q}\mathbf{K}_{\text{I}}^\top  \in \mathbb R^{L \times L_{\text{I}}}$.

Consider the attention score for $i$-th text token, we have:

\begin{equation} \small
    P_{\text{T}_i} = \frac{e^{S_{\text{T}_i}}}
    {
    \sum_{i=1}^{L_{\text{T}}} e^{S_{\text{T}_i}}  + \sum_{j=1}^{L_{\text{I}}} e^{S_{\text{I}_j}
    }
    }
    \label{eq:score_text}
\end{equation}

where $S_{\text{T}_j}$ is the $j$-th column of $S_{\text{T}}$ and $S_{\text{I}_j}$ is the $j$-th column of $S_{\text{I}}$. Then we accumulate the attention scores for all text tokens as follows:

\begin{equation} \small
    \sum_i P_{\text{T}_i} = \frac{\sum_{i=1}^{L_{\text{T}}}e^{S_{\text{T}_i}}}
    {
    \sum_{i=1}^{L_{\text{T}}} e^{S_{\text{T}_i}}  + \sum_{j=1}^{L_{\text{I}}} e^{S_{\text{I}_j}
    }
    }
    \label{eq:sum_score_text}
\end{equation}

Since $L_{\text{I}}$ grows as image resolution grows and $L_{\text{T}}$ remains constant, we can observe that the attention score to text tokens decreases accordingly. Therefore, the attention assigned to each text token decreases.

Although lowering the temperature of softmax can sharpen the distribution, it only enhances tokens that received more attention initially. In other words, it cannot enhance most of the text tokens, especially those not directly related to the main subject.

Heat maps in Figure~\ref{fig:text_decay} demonstrate the accumulated attention scores for all text tokens. We compare the generation at training resolution and the generation at a higher resolution with direct extrapolation, NTK-aware interpolation, NTK-by-parts interpolation and YaRN. We can observe that when generating high-resolution images, the first three methods have very dark heatmaps, and their results suffer from the subject vanishing issue. YaRN recovers brightness in the main subject areas of the heatmap, but still loses attention to text tokens in surrounding areas. While its results recover the subject, other information in the prompt (such as background and image style) is lost.

\noindent \textbf{Spectral Progression of the Diffusion Process.}
\citet{rissanen2022generative, hoogeboom2023simple} have shown the spectral progression of the diffusion process, where the models focus on global structures (low-frequency) in early sampling steps and generate local details (high-frequency) in late steps. This implies that extrapolation methods should also adapt to this pattern.

Based on this observation, \citet{zhuo2024lumina} proposed time-aware scaled RoPE, which uses PI in the early sampling steps for global structure and gradually shifting to NTK-Aware interpolation for local details. \citet{issachar2025dype} further analyzed the evolution of frequency modes in the diffusion process, and introduced explicit time dependence into the formulate of PI, NTK-Aware Interpolation and YaRN. However, both of them are essentially interpolation methods that consider the current sampling step, and they do not dynamically adjust the distribution of attention scores.

\subsection{Text Anchoring}
To address the issue of text decay, we propose a new approach called Text Anchoring to preserve the influence of text tokens during high-resolution generating.

Specifically, we introduce a bias $\beta > 0$ added to the original attention logits to text tokens $\mathbf{S}_\text{T} = \mathbf{Q}\mathbf{K}_{\text{T}}^\top \in \mathbb R^{L \times L_{\text{T}}}$, yielding the corrected logits $\mathbf{S}_{\text{T}}'$:

\begin{equation} \small
    \mathbf{S}_{\text{T}}' =  \mathbf{S}_\text{T} + \beta
    \label{eq:add_bias}
\end{equation}

Given the shift invariance of softmax, adding a bias instead of multiplying a factor can better preserve the relative probability distribution among text tokens. Furthermore, multiplicative enhancement requires additional consideration of negative logits to avoid mistakenly weakening attention, which requires implementing a new kernel. In contrast, additive enhancement only requires passing an attention mask.

We further consider the value of bias $\beta$. Instead of setting it as a constant hyperparameter, we proposed a method adapted to the extrapolation scale. Let the number of pixels in the target images be $\lambda$ times that of the training image, the sum value for image tokens in Eq.~\eqref{eq:sum_score_text}'s denominator can be presumably approximated as $\lambda$ times the original value. Therefore, we expect the sum value for text tokens should also become $\lambda$ times the original value after adding the bias:

\begin{equation} \small 
    \sum_{i=1}^{L_{\text{T}}}e^{S_{\text{T}_i}+\beta}=\lambda \cdot \sum_{i=1}^{L_{\text{T}}}e^{S_{\text{T}_i}}
    \label{eq:derive_beta_0}
\end{equation}

Then we have:

\begin{equation} \small
    e^{x+\beta}=\lambda e^{x}
    \label{eq:derive_beta_1}
\end{equation}

Therefore, we can determine that the bias $\beta$ is:

\begin{equation} \small 
    \beta = \ln(\lambda)
    \label{eq:derive_beta_2}
\end{equation}

If we apply the scaling factor $s$ to both height and width, then we should have $\lambda = s^2$, and bias:

\begin{equation} \small
    \beta = \ln(s^2) = 2 \ln(s)
    \label{eq:derive_beta_3}
\end{equation}

After aligning the influence of text tokens and image tokens, we still need to sharpen the attention distribution with lowering the temperature $\tau$. Thus we have the final attention score: 

\begin{equation} \small 
    \mathbf{P} = \text{Softmax}\left(\frac{ \text{Concat}(\mathbf{S}_\text{T}+\beta, \mathbf{S}_\text{I})}{\tau\sqrt{d}}\right)
    \label{eq:beta_and_temperature}
\end{equation}

where temperature value $\tau$ is same as YaRN's in Eq.~\eqref{eq:yarn_temperature}.

\subsection{Dynamic Temperature Control}
Although the method in section 3.2 restored most of the information in the prompt, it primarily improved the quality of global structure. The local details instead have significant artifacts including speckles and irregular grids due to the sharpening of the attention score's distribution. In other words, while the high-frequency details do require lowering the temperature to prevent blurring, the temperature should not be set too low.

Since diffusion models focus on low-frequency structure in early steps and generate high-frequency details in late steps, we can dynamically control the temperature $\tau$, allowing it to gradually increase during the denoising process. We normalized the time $t$ to the interval $[0, 1]$, which decrease from $1$ to $0$ during the denoising process, then we have $\tau(t)$ as:

\begin{equation} \small 
    \tau(t) = \tau_{\text{max}} - (\tau_{\text{max}} - \tau_{\text{min}}) \cdot t
    \label{eq:heating}
\end{equation}

where $\tau_{\text{max}}$ and $\tau_{\text{min}}$ are constant hyperparameters and satisfy $\tau_{\text{max}} > \tau_{\text{min}}$. In our experiments, we set $\tau_{\text{max}}$ to $1.0$ and $\tau_{\text{min}}$ to the temperature in Eq.~\eqref{eq:yarn_temperature} used by YaRN.

\begin{figure*}[ht!]
    \centering
    \includegraphics[width=0.9\textwidth]{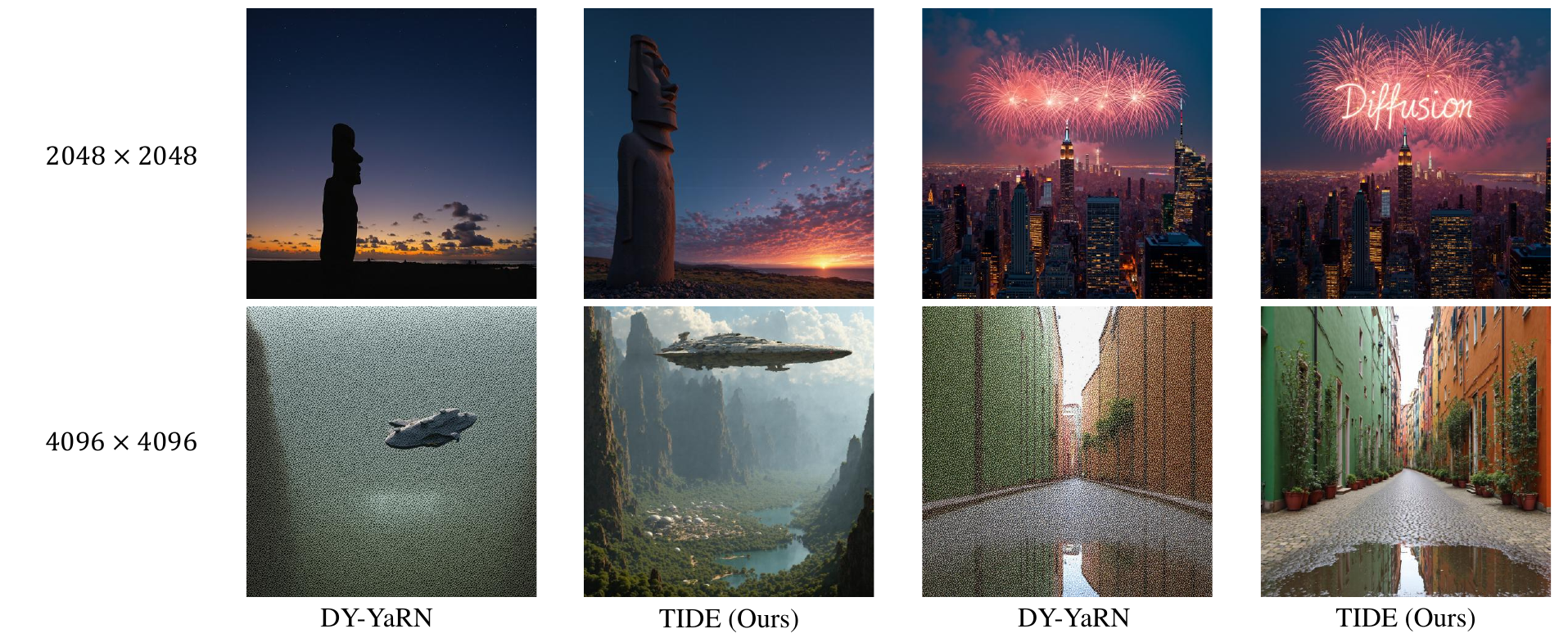}
    \caption{\textbf{Qualitative comparison with baseline method.} We compare TIDE against baseline methods. 
    The input prompts correspond to the rows: 
    \textbf{2K Prompt 1:} ``\textit{A Moai statue gazes upward against a starry night sky, with a colorful sunset illuminating the horizon.}''; 
    \textbf{2K Prompt 2:} ``\textit{New York Skyline with 'Diffusion' written with fireworks on the sky.}''; 
    \textbf{4K Prompt 1:} ``\textit{A futuristic spaceship flies over a lush landscape featuring rocky cliffs and a settlement with domed buildings near a serene body of water surrounded by greenery.}''; 
    \textbf{4K Prompt 2:} ``\textit{Narrow cobblestone alley featuring colorful buildings with green and orange facades, potted plants lining the street, and a puddle reflecting the scene.}''.
    }
    \label{fig:result}
\end{figure*}

Furthermore, DyPE shows that the low-frequency modes evolve fast and converge quickly, while the high-frequency modes evolve gradually and nearly linearly. Inspired by this spectral evolution result, we further adjust the temperature based on different frequencies $f$ of PE. We introduce an exponent $\alpha(f)$ to control the convexity of the curve.

\begin{equation} \small 
    \tau(t, f) =  \tau_{\max} - (\tau_{\max}-\tau_{\min}) \cdot t^{\alpha(f)}
    \label{eq:dyheating_0}
\end{equation}

where $\alpha(f)$ can be defined as:

\begin{equation} \small 
    \alpha(f)=\alpha_{\text{low}} + (\alpha_{\text{high}}-\alpha_{low}) \cdot f
    \label{eq:dyheating_1}
\end{equation}

The $\alpha_{\text{low}}$ and $\alpha_{\text{high}}$ are constant hyperparameters corresponding to the lowest frequency and highest frequency respectively, and satisfy $\alpha_{\text{low}} > \alpha_{\text{high}}$. In our experiments, we set $\alpha_{\text{low}}$ to $0.6$ and $\alpha_{\text{high}}$ to $0.2$.

\section{Experiments}

\subsection{Experimental Setup}

\noindent \textbf{Settings.} We apply TIDE on FLUX.1-dev \citep{flux2024}, an advanced open-source model based on DiT architecture with default generation resolution as $1024 \times 1024$. We experiment with resolution up to $4096 \times 4096$. The sampling steps are set to $28$, and the guidance scale is set to $3.5$. For our dynamic temperature control, we apply $\alpha_{\text{low}}=0.6$ and $\alpha_{\text{high}}=0.2$. All experiments are conducted on NVIDIA RTX 6000 Ada GPUs.

\noindent \textbf{Baselines.} We compare our method with three baselines: (1) Direct Extrapolation; (2) YaRN \citep{peng2023yarn}; (3) Dy-YaRN \citep{issachar2025dype}.

\noindent \textbf{Metrics.} We choose Fréchet Inception Distance \citep{heusel2017gans} (FID) and Kernel Inception Distance \citep{binkowski2018demystifying} (KID), CLIP Score \citep{radford2021learning}, ImageReward \citep{xu2023imagereward} and Aesthetic-Score-Predictor \citep{schuhmann2022laion}. Both FID and KID measure the similarity between the ground images and the generated images. CLIP Score represents the prompt-following capability, ImageReward represents human preference alignment, and Aesthetic Score represents human aesthetic judgments. Note that since the count of images is small, KID, which is unbiased, has better reliability than FID.

\subsection{Qualitative Results}
\label{sec:qualitative_results}
Figure~\ref{fig:result} provides qualitative comparisons at different resolutions with prompts from DrawBench and Aesthetic-4K.
With prompt 1 at $2048 \times 2048$ resolution, our images exhibit superior aesthetic quality. The colors of sunsets and the details of statues appear richer, while the composition has enhanced visual appeal. Our method also preserves the full information from the prompt 2 at $2048 \times 2048$ resolution, especially the word `Diffusion'. As resolution increases to $4096 \times 4096$, our method maintains quality without degradation in content richness or detail fidelity. More quantitative results comparisons are provided in Appendix~\ref{sec:qualitative}.

\subsection{Quantitative Results}

\begin{table*}[t]
    \centering
    \caption{\textbf{Quantitative comparison with baselines.} We compare our TIDE with FLUX, YaRN, and Dy-YaRN on DrawBench and Aesthetic-4K. All methods are built on FLUX. Best results are \textbf{bolded}.}
    \label{tab:main_results}
    
    \resizebox{.9\linewidth}{!}{
        \begin{tabular}{c l ccc ccccc}
            \toprule
            \multirow{2}{*}{\textbf{Resolution}} & \multirow{2}{*}{\textbf{Method}} & \multicolumn{3}{c}{\textbf{DrawBench}} & \multicolumn{5}{c}{\textbf{Aesthetic-4K}} \\
            \cmidrule(lr){3-5} \cmidrule(lr){6-10}
            
             & & CLIP$\uparrow$ & IR$\uparrow$ & Aesth$\uparrow$ & FID$\downarrow$ & KID$\downarrow$ & CLIP$\uparrow$ & IR$\uparrow$ & Aesth$\uparrow$ \\
            \midrule

            \multirow{4}{*}{$2048 \times 2048$} 
              & FLUX    & 26.63 &  0.46 & 5.53 & 162.59 & 0.0087 & 27.87 & 1.02 & 6.54 \\
              
              & YaRN    & 26.76 &  0.66 & 5.76 & \textbf{141.93} & \textbf{0.0045} & 28.07 & 1.14 & 6.59 \\
              
              & Dy-YaRN & 26.52 & 0.72 & 5.74 & 161.32 & 0.0068 & 28.09 & 1.10 & 6.62 \\
              
              & \textbf{TIDE (Ours)} & \textbf{27.34} & \textbf{0.83} & \textbf{5.77}  & 159.35 & 0.0066 & \textbf{28.60} & \textbf{1.21} & \textbf{6.68} \\
            
            \midrule
            
            \multirow{4}{*}{$4096 \times 4096$} 
              & FLUX    & 15.16 &  -2.19 & 2.92 & 341.79 & 0.1952 & 13.46 & -2.25 & 2.58 \\
              
              & YaRN    & 23.25 &  -0.83 & 4.39 & 211.48 & 0.0494 & 23.86 & -0.24 & 5.75 \\
              
              & Dy-YaRN & 23.21 & -0.84 & 4.39 & 211.50 & 0.0496 & 23.90 & -0.24 & 5.75 \\
              
              & \textbf{TIDE (Ours)} & \textbf{26.13} & \textbf{0.18} & \textbf{5.34} & \textbf{158.39} & \textbf{0.0081} & \textbf{27.33} & \textbf{0.75} & \textbf{6.44} \\
            
            \bottomrule
        \end{tabular}
    }
\end{table*}
To evaluate the quantitative results, we use all 200 text prompts from DrawBench \citep{saharia2022photorealistic} and all 195 image-prompt pairs from the 4K-resolution subset of Aesthetic-4K \citep{zhang2025diffusion} to generate images. 
The results are summarized in Table~\ref{tab:main_results}. At $2048 \times 2048$ resolution, our method achieves the best results across most metrics. Based on the quantitative results from Section~\ref{sec:qualitative_results}, the baseline methods can still generate image with acceptable quality. As resolution increases to $4096 \times 4096$, our method demonstrates a more pronounced advantage. Direct extrapolation suffers from severe decline in all metrics, corresponding to the subject vanishing issue illustrated in Figure~\ref{fig:subject_vanishing}. While YaRN and Dy-YaRN exhibit improvements over raw FLUX, they still display considerable gaps compared to our method. These results demonstrate our advantages in ultra-high-resolution text-to-image synthesis.

\subsection{User Study}
We conduct a User Study with researchers familiar with generative AI models. We randomly selected 20 prompts from Aesthetic-4K to generate images with Dy-YaRN and our TIDE at $4096 \times 4096$ resolution. For each prompt, participants were required to rate images on a scale from 1 to 5 at three levels: (i) Alignment with the text. (ii) Structural integrity and proportions. (iii) Refined textures and details. As illustrated in Table~\ref{tab:user_study}, TIDE achieves much better scores over three aspects.

\begin{table}[t]
    \centering

    \caption{\textbf{User study results.} We conduct a subjective evaluation comparing TIDE and Dy-YaRN on 20 randomly selected prompts. Participants rated images on a scale of 1 to 5 (higher is better). We report the mean score for each aspect.}
    \label{tab:user_study}

    \resizebox{0.8\linewidth}{!}{
        \begin{tabular}{l c c}
            \toprule
            \textbf{Method} & \textbf{TIDE (Ours)} & \textbf{Dy-YaRN} \\
            \midrule
            Text Alignment   & \textbf{4.57} & 3.60 \\
            Global Structure & \textbf{4.33} & 3.21 \\
            Texture Quality  & \textbf{4.46} & 2.48 \\
            \bottomrule
        \end{tabular}
    }
\end{table}

\subsection{Ablation study}
\label{sec:main_ablation}
We conduct the ablation study to validate the effectiveness of our two core components: Text Anchoring (TA) and Dynamic Temperature Control (DTC). We apply Dy-YaRN \citep{issachar2025dype} on FLUX as our baseline. We randomly selected 50 prompts from Aesthetic-4K \citep{zhang2025diffusion} to guide image generation at $4096 \times 4096$ resolution across four configurations: (1) Baseline (w/o TA and DTC), (2) TA only, (3) DTC only, and (4) Full Method (w/ TA and DTC). 

\begin{table}[t]
    \centering
    \caption{\textbf{Component-wise ablation study.} We investigate the impact of Text Anchoring and Dynamic Temperature Control on Asethetic-4K. The baseline model is Dy-YaRN. Best results are \textbf{bolded}.}
    \label{tab:ablation_components}
    
    \resizebox{.8\linewidth}{!}{
        \begin{tabular}{c ccc}
            \toprule
            Configurations & CLIP$\uparrow$ & IR$\uparrow$ & Aesth$\uparrow$ \\
            \midrule

              Baseline & 23.60 & -0.44 & 5.64 \\
            
            TA Only & 26.60 & 0.66 & \textbf{6.54} \\
            
            DTC Only & 23.28 & -0.30 & 5.65 \\
            
            Full Method & \textbf{26.76} & \textbf{0.86} & 6.46 \\
            
            \bottomrule
        \end{tabular}
    }
\end{table}

The quantitative results are summarized in Table~\ref{tab:ablation_components}. Regardless of whether DTC is enabled, enabling TA has a significant positive effect on metrics. However, Enabling DTC may result in a decline in certain metrics. Despite this, qualitative experiment reveals that the generated images tend to suffer from severe artifacts without DTC. Further qualitative results and parameter analysis supporting this design choice can be found in Appendix~\ref{sec:ablation} and Figure~\ref{fig:why_heating}.

\section{Conclusion}

We presented TIDE, a training-free framework that enables pre-trained diffusion transformers to generate images with higher resolution without additional sampling overhead. We identified the issue of prompt information loss and proposed the text anchoring method to improve the global structure. We also introduced the dynamic temperature control mechanism  to eliminate local artifacts. Extensive experiments demonstrated that TIDE significantly improves resolution extrapolation performance and has excellent compatibility.

As future work, we will focus on the TIDE framework guided by real-time data such as attention entropy, which involves high-performance kernel design and system-level optimization. We also hope to find similar insights from other generative tasks, such as image-to-image generation and video generation, to improve their extrapolation quality.


\section*{Impact Statement}
This paper presents work whose goal is to advance the field of machine learning. There are many potential societal consequences of our work, none of which we feel must be specifically highlighted here.

\bibliography{references}
\bibliographystyle{icml2026}

\newpage
\appendix
\onecolumn

\section{Time-Shifting Schedule Setting}

In models like Stable Diffusion 3 \citep{esser2024scaling} and FLUX.1 \cite{flux2024}, the discrete time-steps are mapped to a continuous time domain $t \in [0, 1]$ via a shifting function. As the denoising process progresses, $t$ gradually decreases from $1$ to $0$. A shift parameter, denoted as $\mu$, controls the density of time-steps towards the noise (start) or the data (end). A higher $\mu$ biases the sampling trajectory to spend more steps in the high-noise region (near $t=1$). 

FLUX.1 adopts a dynamic shifting strategy, where $\mu$ scales linearly with the number of image tokens $L_{\text{I}}$, defined by anchor points at $L_{\text{I}} = 256$ ($\mu=0.5$) and $L_{\text{I}}=4096$ ($\mu=1.15$). While this design aims to enhance global structure generation at higher resolutions, we observe that the default linear scaling becomes problematic for extreme high resolutions. For instance, at $4096 \times 4096$ resolution, where $L_{\text{I}}=65536$, the linearly extrapolated $\mu$ can reach approximately $11.5$. With a standard 28-step scheduler, this causes the effective $t$ to remain near $1.0$ for the first 27 steps, forcing the model to condense the entire denoising process into the final step. This results in severe degradation of generation quality.

We note that \citet{issachar2025dype} attempted to mitigate this by clamping $\mu$ at $1.15$ for resolutions beyond $1024^2$ in their code. However, this ignores the increased complexity of higher-resolution latents. To address this, we use a \textbf{logarithmic time-shifting schedule} for $\mu$ with respect to $L_{\text{I}}$: 

\begin{equation}
    \mu =  k \cdot \ln (L_{\text{I}}) + b 
    \label{eq:shift_log}
\end{equation}

where $k$ and $b$ are determined by default anchor points at $L_{\text{I}} = 256$ ($\mu=0.5$) and $L_{\text{I}}=4096$ ($\mu=1.15$).

This approach maintains a smoother increase in shift values, preventing the issue of default setting while still adapting to resolution changes. To ensure a fair comparison, all experiments in this paper (including baseline evaluations) have used this logarithmic time-shifting schedule.
\section{Additional Ablation Studies}
\label{sec:ablation}
In this section, we conduct further ablation studies on hyperparameters of our two core components: bias value for Text Anchoring (TA) and $\alpha_{\text{low}}$ and $\alpha_{\text{high}}$ for Dynamic Temperature Control (DTC). We use Dy-YaRN \citep{issachar2025dype} on FLUX as reference baseline. The experimental setting is same as Section~\ref{sec:main_ablation} unless otherwise specified.

\subsection{Ablation study on Text Anchoring}
\begin{figure}[ht]
    \centering

    \includegraphics[width=0.5\linewidth]{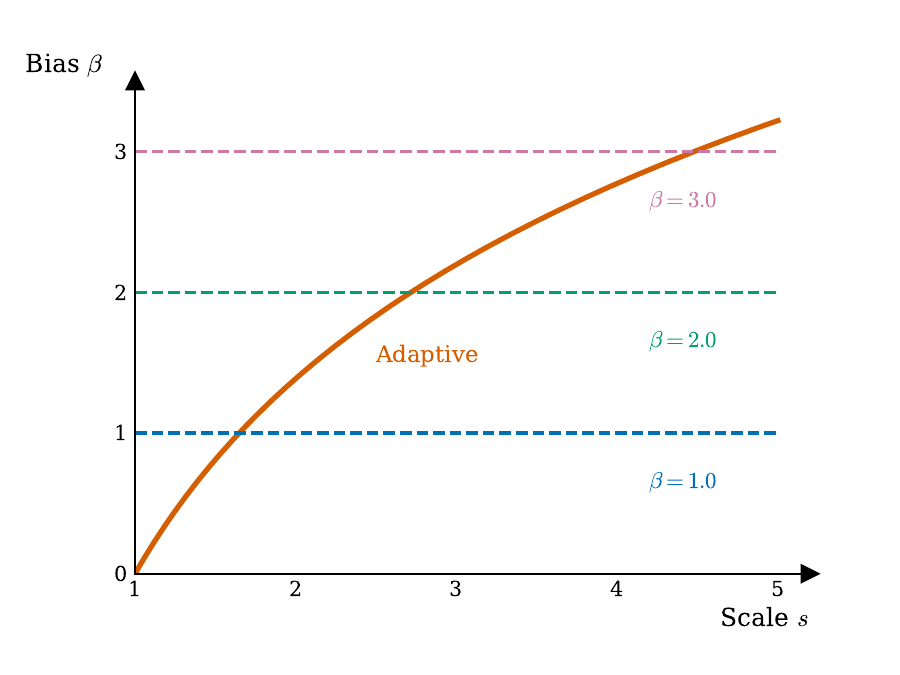}

    \caption{\textbf{Scale-adaptive bias.} Our adaptive bias increases logarithmically with the extrapolation scale.}
    \label{fig:ta_bias}
\end{figure}
\begin{table*}[!ht]
    \centering
    \caption{\textbf{Ablation study on Text Anchoring bias across different resolutions.} We evaluate the impact of bias strategies on image generation quality at $2048^2$, $3072^2$, and $4096^2$ resolutions. Dy-YaRN is adopted as the baseline. The best results for each resolution are \textbf{bolded}, and the second-best results are \underline{underlined}.}
    \label{tab:ablation_bias}
    
    \resizebox{\textwidth}{!}{
        \begin{tabular}{lc ccc ccc ccc}
            \toprule
            \multirow{2}{*}{\textbf{Method}} & \multirow{2}{*}{\textbf{Bias Scale}} & \multicolumn{3}{c}{\textbf{2048 $\times$ 2048}} & \multicolumn{3}{c}{\textbf{3072 $\times$ 3072}} & \multicolumn{3}{c}{\textbf{4096 $\times$ 4096}} \\
            \cmidrule(lr){3-5} \cmidrule(lr){6-8} \cmidrule(lr){9-11}
             & & CLIP$\uparrow$ & IR$\uparrow$ & Aesth$\uparrow$ & CLIP$\uparrow$ & IR$\uparrow$ & Aesth$\uparrow$ & CLIP$\uparrow$ & IR$\uparrow$ & Aesth$\uparrow$ \\
            \midrule
            
            Dy-YaRN & - & 29.10 & 1.12 & 6.83 & 26.87 & 0.69 & 6.56 & 23.60 & -0.44 & 5.64 \\
            
            \cmidrule(lr){1-11}
            
            \multirow{4}{*}{TIDE (Ours)} 
             & 1.0 & \textbf{29.16} & 1.15 & \textbf{6.86} & 27.81 & 0.94 & 6.78 & 26.05 & 0.54 & 6.26 \\
             & 2.0 & 28.87 & \textbf{1.19} & 6.80 & 28.08 & 1.08 & 6.76 & \underline{26.91} & \underline{0.83} & \textbf{6.47} \\
             & 3.0 & 28.42 & 1.07 & 6.84 & 28.14 & 1.03 & 6.77 & \textbf{27.25} & 0.71 & 6.41 \\
             \addlinespace[2pt]
             & \textbf{Adaptive} & \underline{29.12} & \underline{1.17} & \underline{6.85} & \textbf{28.28} & \textbf{1.09} & \textbf{6.78} & 26.76 & \textbf{0.86} & \underline{6.46} \\
            
            \bottomrule
        \end{tabular}
    }
\end{table*}

The bias parameter $\beta$ in Text Anchoring governs the strength of the token influence. We compare fixed bias values ($\beta \in \{1.0, 2.0, 3.0\}$) against our proposed scale-adaptive design in Eq.~\eqref{eq:derive_beta_3}. Figure~\ref{fig:ta_bias} demonstrates how bias value changes with extrapolation scale $s$.

Table~\ref{tab:ablation_bias} reports the performance across various resolutions. While a fixed bias (e.g., $\beta=1.0$) may perform marginally better at specific resolution, it lacks generalization capabilities for different resolutions. In contrast, our Adaptive scheme consistently yields optimal or near-optimal performance across the entire resolution spectrum. This confirms that the optimal bias is intrinsically related to the length of the token sequence.

\subsection{Ablation study on Dynamic Temperature Control}
\label{sec:ablation_dtc}
\begin{figure}[ht]
    \centering

    \includegraphics[width=\linewidth]{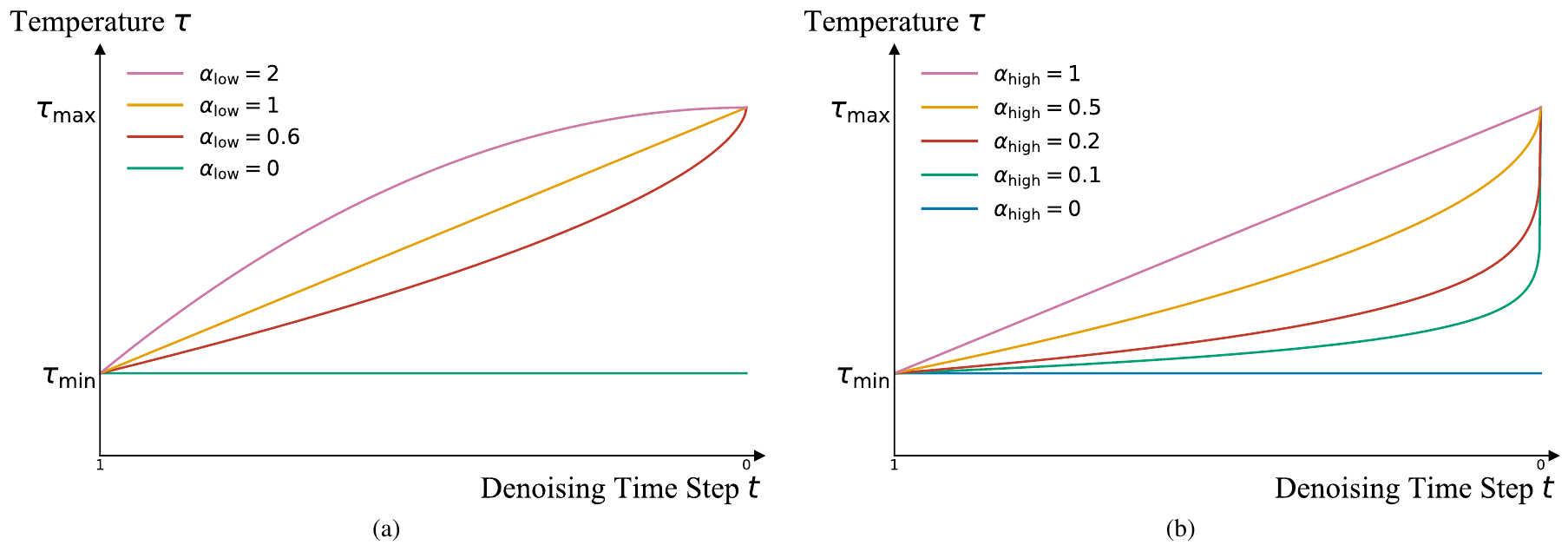}

    \caption{\textbf{Temperature evolution patterns with different $\alpha$.} (a) demonstrates different patterns for component with lowest frequency of positional embeddings. (b) demonstrates different patterns for component with highest frequency of positional embeddings.}
    \label{fig:dtc_alpha}
\end{figure}
\begin{table}[!ht]
    \centering
    \caption{\textbf{Hyperparameter sensitivity of Dynamic Temperature Control.} We evaluate the impact of low-frequency ($\alpha_{\text{low}}$) and high-frequency ($\alpha_{\text{high}}$) exponents. Dy-YaRN serves as the baseline. The overall best results are \textbf{bolded}.}
    \label{tab:ablation_temperature}
    \begin{tabular}{lc c ccc}
        \toprule
        \textbf{Method} & \textbf{$\alpha_{\text{low}}$} & \textbf{$\alpha_{\text{high}}$} & CLIP$\uparrow$ & IR$\uparrow$ & Aesth$\uparrow$ \\
        \midrule
        
        Dy-YaRN & - & - & 23.60 & -0.44 & 5.64 \\
        
        \cmidrule(lr){1-6}
        
        \multirow{9}{*}{TIDE} 
         & 0.0 & 0.0 & 26.60 & 0.66 & 6.54 \\
         
         \cmidrule(lr){2-6}
         
         & \multirow{4}{*}{0.6} & 0.1 & \textbf{27.25} & 0.84 & 6.49 \\
         &                      & 0.2 & 26.76 & \textbf{0.86} & 6.46 \\
         &                      & 0.5 & 27.09 & 0.84 & \textbf{6.52} \\
         &                      & 1.0 & 26.03 & 0.53 & 6.19 \\
         
         \cmidrule(lr){2-6} 
         
         & \multirow{2}{*}{1.0} & 0.2 & 27.19 & 0.84 & 6.49 \\
         &                      & 1.0 & 27.03 & 0.76 & 6.46 \\
         
         \cmidrule(lr){2-6}
         
         & \multirow{2}{*}{2.0} & 0.2 & 26.99 & 0.78 & 6.38 \\
         &                      & 1.0 & 26.97 & 0.73 & 6.37 \\
        
        \bottomrule
    \end{tabular}
\end{table}

The DTC component is parameterized by $\alpha_{low}$ and $\alpha_{high}$ in Eq.~\eqref{eq:dyheating_0} and Eq.~\eqref{eq:dyheating_1} corresponding to the lowest frequency and highest frequency, respectively. Setting $\alpha_{low}=\alpha_{high}=0$ is equivalent to disabling DTC. Figure~\ref{fig:dtc_alpha} illustrates how different values of $\alpha$ influence the patterns of temperature evolution.

First, we fix $\alpha_{\text{high}}=0.2$ and vary $\alpha_{\text{low}}\in \{0.6, 1, 2\}$. Based on qualitative and quantitative evaluation, we fix $\alpha_{\text{low}} = 0.6$ and vary $\alpha_{\text{high}} \in \{0.1,0.2,0.5,1.0\}$. Quantitative results are summarized in Table~\ref{tab:ablation_temperature}. 

\begin{figure}[!ht]
    \centering

    \includegraphics[width=0.8\linewidth]{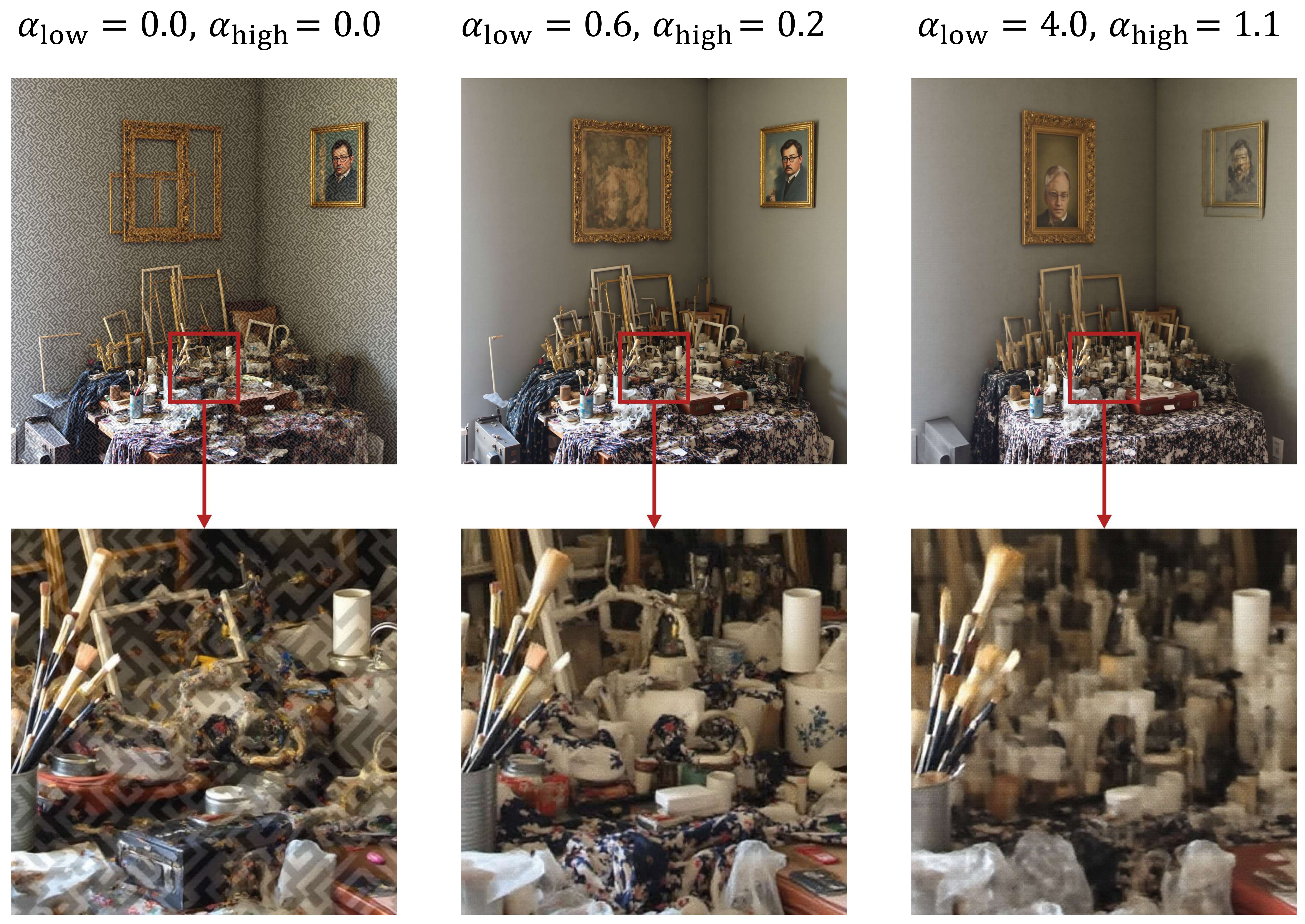}

    \caption{\textbf{Qualitative comparison at different $\alpha_{\text{low}}$ and $\alpha_{\text{high}}$.} The image on the left (with $\alpha_{\text{low}}=0.0$ and $\alpha_{\text{high}}=0.0$) suffers from severe artifacts, while the image on the right shows blurry details. The image in the middle has the best quality. }
    \label{fig:why_heating}
\end{figure}

It is worth noting that no single combination of parameters achieves optimal results across all parameters. As illustrated in Figure~\ref{fig:why_heating}, severe artifacts can be observed when $\alpha_{\text{low}}$ and $\alpha_{\text{high}}$ are set to zero, meanwhile excessively high values of $\alpha_{\text{low}}$ and $\alpha_{\text{high}}$ cause image blurring. Based on both quantitative results and qualitative observations, we ultimately set $\alpha_{\text{low}}=0.6$ and $\alpha_{\text{high}}=0.2$.

\section{Additional Qualitative Results}
\label{sec:qualitative}

\begin{figure}[ht]
    \centering

    \includegraphics[width=0.4\linewidth]{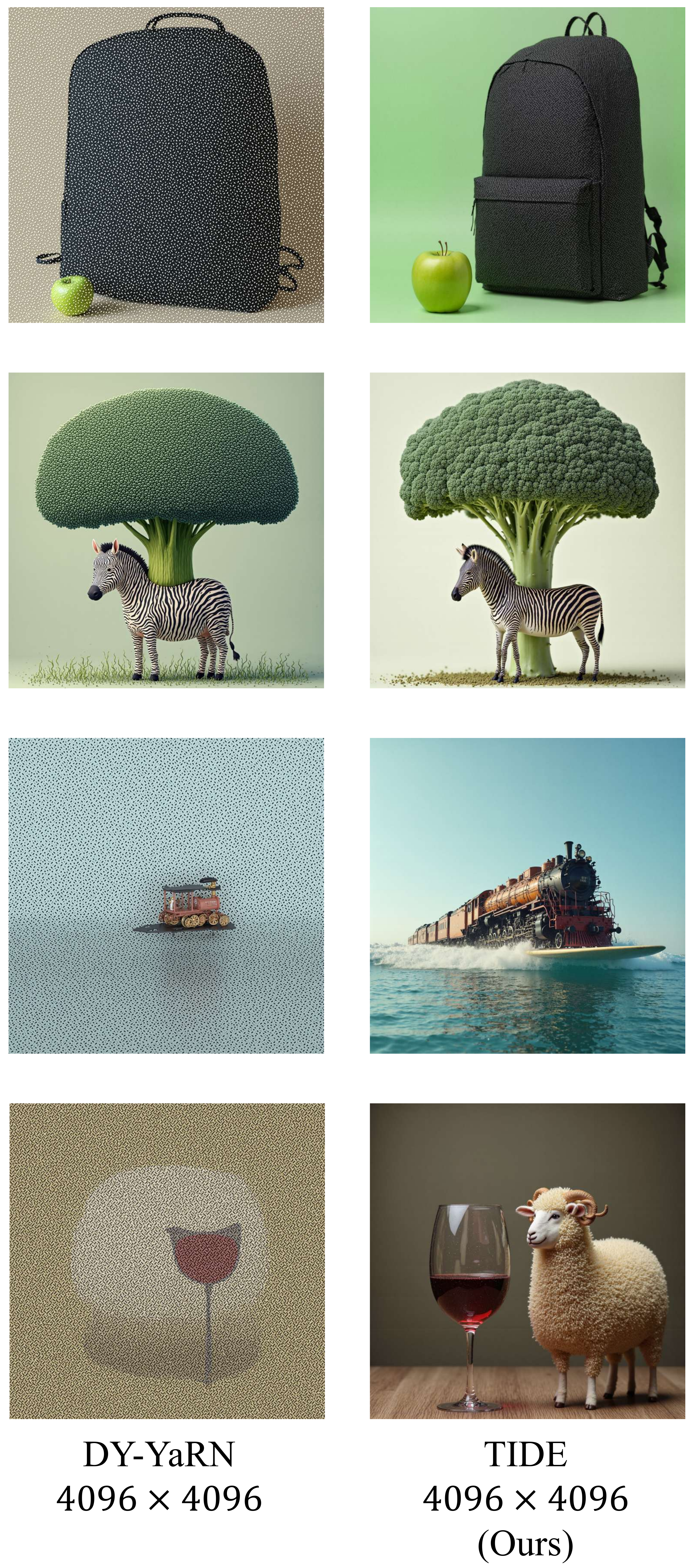}

    \caption{\textbf{Qualitative comparison for high-resolution text-to-image generation on DrawBench.} Our method outperforms baselines across by delivering high visual quality.}
    \label{fig:result_drawbench}
\end{figure}
\begin{figure}[ht]
    \centering

    \includegraphics[width=0.4\linewidth]{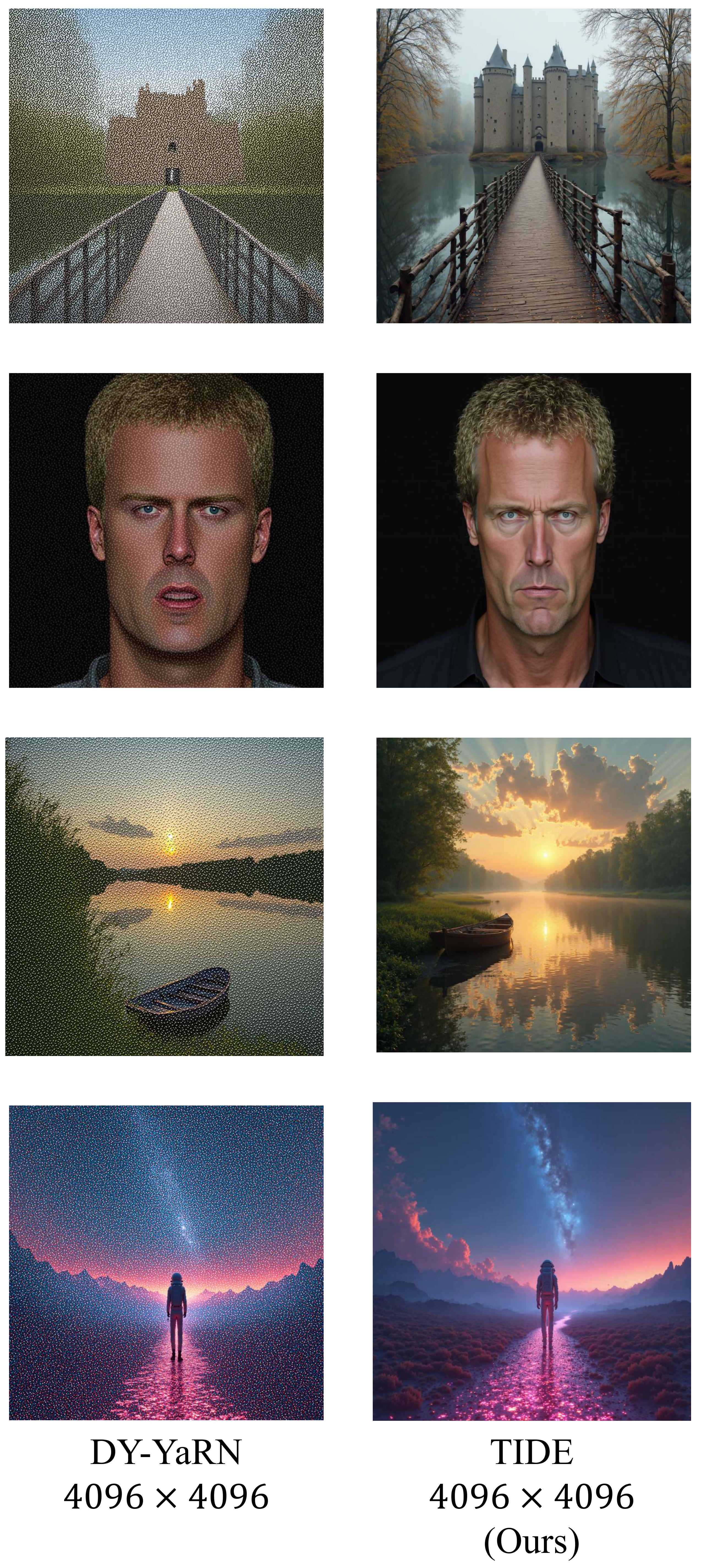}

    \caption{\textbf{Qualitative comparison for high-resolution text-to-image generation on Aesthetic-4K.} Our method outperforms baselines across by delivering high visual quality.}
    \label{fig:result_aesthetic}
\end{figure}

We present further qualitative results, comparing our approach to existing baselines. Figure~\ref{fig:result_drawbench} and Figure~\ref{fig:result_aesthetic} show the generated results on DrawBench \citep{saharia2022photorealistic} and Aesthetic-4K \citep{zhang2025diffusion}, respectively. 

\begin{figure}[ht]
    \centering

    \includegraphics[width=0.8\linewidth]{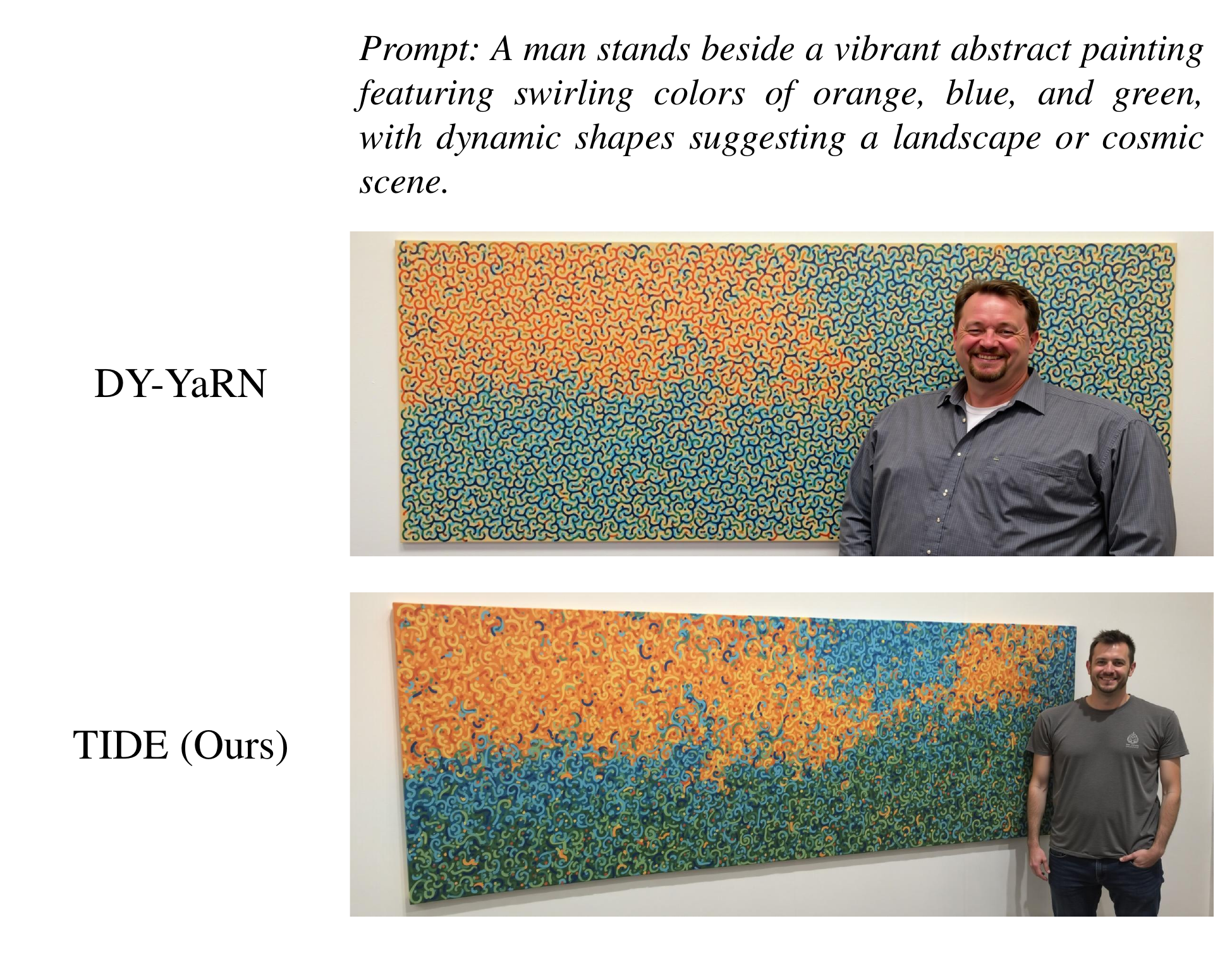}

    \caption{Qualitative comparison of panoramic generation at $4096 \times 1536$ resolution. }
    \label{fig:panoramic_a}
\end{figure}
\begin{figure}[ht]
    \centering

    \includegraphics[width=0.8\linewidth]{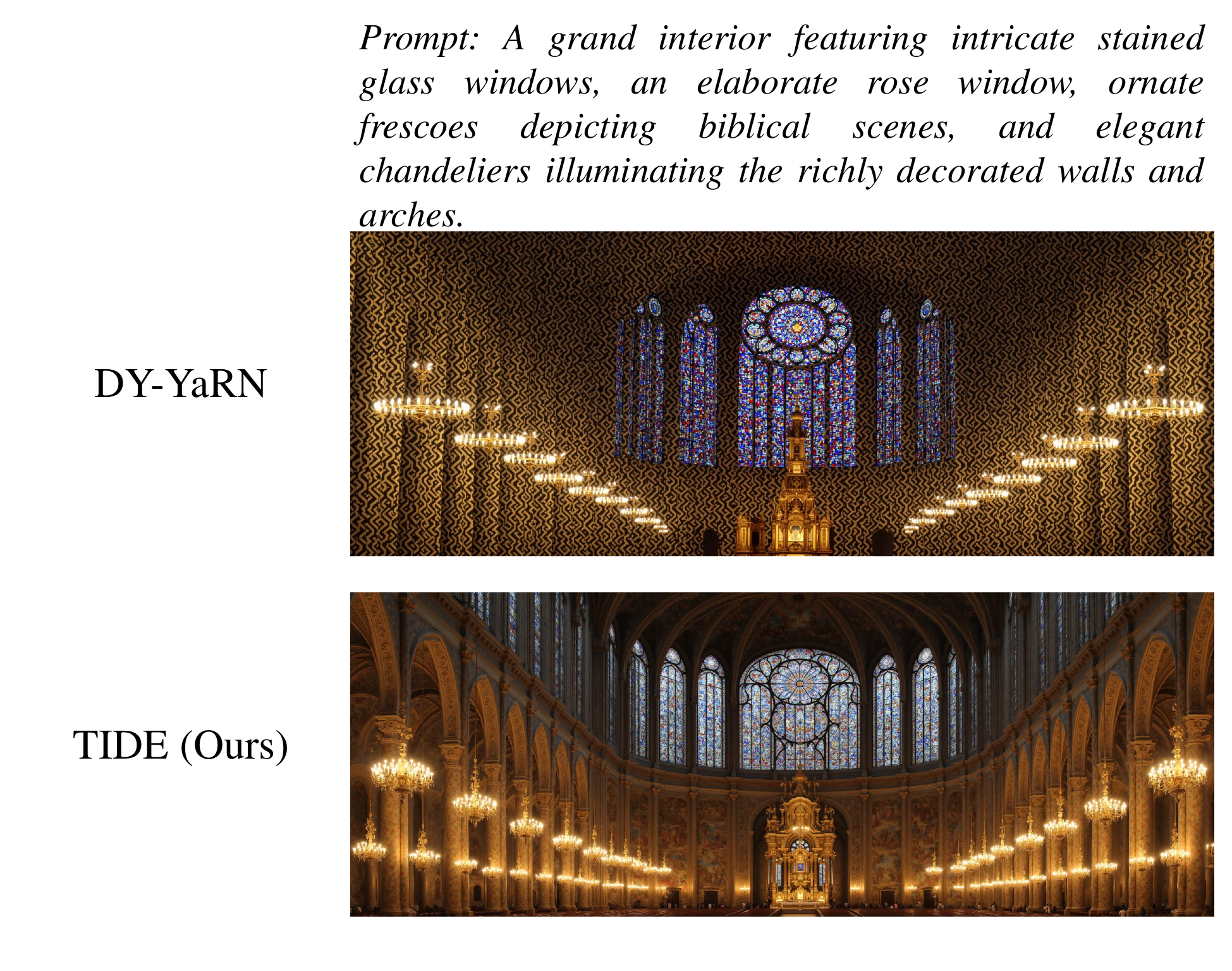}

    \caption{Qualitative comparison of panoramic generation at $4096 \times 1536$ resolution. }
    \label{fig:panoramic_b}
\end{figure}

We also present additional comparison on panoramic image generation at $4096 \times 1536$ resolution. As illustrated in Figure~\ref{fig:panoramic_a}, compared to YaRN, our method presents accurate human proportions. Benefiting from Text Anchoring component, our approach incorporates more information about the human figures. Figure~\ref{fig:panoramic_b} demonstrates that Dy-YaRN suffers from insufficient information diversity and severe artifacts, while our method effectively improves both global and local quality.

\end{document}